\documentclass{article}

\usepackage[final]{neurips}

\usepackage[utf8]{inputenc} 
\usepackage[T1]{fontenc}    
\usepackage{hyperref}       
\usepackage{url}            
\usepackage{booktabs}       
\usepackage{amsfonts}       
\usepackage{nicefrac}       
\usepackage{microtype}      
\usepackage{xcolor}         
\usepackage{color}

\usepackage{graphicx}
\usepackage{enumitem}   

\title{Probabilistic Models for Manufacturing Lead Times}

\author{%
  Recep Yusuf Bekci \\
  McGill University\\
  Montreal, Canada \\
  \texttt{recep.bekci@mail.mcgill.ca} 
   \And
   Yacine Mahdid \\
  Axya \\
  Montreal, Canada \\
  \texttt{yacine@axya.co}
  \And
  Jinling Xing  \\
  Axya\\
  Montreal, Canada \\
  \texttt{jinling@axya.co}
  \And
  Nikita Letov \\
  McGill University\\
  Montreal, Canada \\
  \texttt{nikita.letov@mail.mcgill.ca} 
  \And
  Ying Zhang \\
  McGill University\\
  Montreal, Canada \\
  \texttt{ying.zhang8@mail.mcgill.ca} 
  \And
  Zahid Pasha  \\
  Axya\\
  Montreal, Canada \\
  \texttt{pasha@axya.co}
}

\begin{document}

\maketitle
\begin{abstract}
  In this study, we utilize Gaussian processes, probabilistic neural network, natural gradient boosting, and quantile regression augmented gradient boosting to model lead times of laser manufacturing processes. We introduce probabilistic modelling in the domain and compare the models in terms of different abilities. While providing a comparison between the models in real-life data, our work has many use cases and substantial business value. Our results indicate that all of the models beat the company estimation benchmark that uses domain experience and have good calibration with the empirical frequencies.
\end{abstract}

\section{Introduction}
\label{sec:intro}

Cost estimation has a fundamental function for manufacturing operations. Typically, the total cost of manufacturing a part is a sum of different elements such as material cost, labor cost, engineering cost, and operation costs. In this study, we collaborate with a metal manufacturing firm whose cost structure is the sum of the material cost and the cost of laser operations. In this case, the cost of laser operations is volatile and it is not easy to determine this element before completing the production process, nevertheless, the material costs have a deterministic structure. To determine the laser costs, production lead times (i.e. the compilation times of laser processes) are needed since machining and labor can be represented as time amounts. Moreover, once the lead time is known, it is just a simple calculation to determine the laser costs. Furthermore, production lead times are also critical for production planning and customer satisfaction. Correct information about lead times would increase the accuracy and efficiency of production planning, which naturally extend the benefits through the whole supply chain. As a result of correct production planning, manufacturers can give better deadline promises and build customer satisfaction.

Uncertainty quantification is an important research area in machine learning. In contrast with the deterministic view of machine learning, probabilistic models can be described by the phrase "the models that know what they don't know". Likewise in many processes in the business, laser production lead times also have an uncertain nature because of the complexity of the process. Therefore, the lead time modelling constitutes a natural yet not explored application area for probabilistic models. The variability in laser lead times can be explained up to some extent by the features of the part to be produced using pattern recognition techniques. Previous studies in this area focused on point estimates (see our discussion in Section \ref{sec:literature}). However, quantifying the uncertainty of estimations is also valuable for practical reasons. The source of the uncertainty that we are interested in can be cast into two: \begin{enumerate}[label=\roman*.]

    \item Epistemic uncertainty: uncertainty due to the model

    \item Aleatoric uncertainty: uncertainty that captures noise in the data.

\end{enumerate}

 Our model corpus varied in terms of the ability to capture these uncertainties (see \cite{gal2016uncertainty} for an extensive discussion on this topic).

Probabilistic modelling of lead times also have substantial business value for pricing decisions. The prediction intervals can serve as functional decision tools when quoting for a product. For instance, taking a predetermined confidence level, the decision maker can quote for at the lower bound  in order to be in front of the competition, or otherwise they can quote at the upper bound to make extra profit. Furthermore, for automatic quoting, higher uncertainty can be used as a signal for human intervention requirement. 

In this paper, we present our methodology and results for the data-driven prediction of laser lead times using data from a metal manufacturing firm that operates in Canada. This problem has drawn extensive attention in the literature due to its importance. While the nature of the problem is stochastic and probabilistic modelling of the problem has significant benefits, the majority of the literature is populated by point estimate methods. We introduce probabilistic modelling into the domain while exploring various probabilistic machine learning algorithms. Our results indicate that our models beat linear regression baseline and the company estimation benchmark that uses domain experience. Furthermore, the probabilistic models have good calibration with the empirical frequencies which validates their capturing abilities of the variability.

\section{Related Work}
\label{sec:literature}

Predictive modelling of manufacturing lead times is a novel research stream aligned with the increased data availability and recent developments in artificial intelligence and machine learning. One of the most recent works in cost estimation is \cite{loyer2016comparison}. Their approach is a case study of a cost estimation problem in jet engine components. They applied various techniques including multiple linear regression and artificial neural networks, support vector regression and gradient boosted trees. They use six main features in their analysis: the part category, two features that describe the part geometry, two for material properties (machinability, cost rate) and the last one is the production volume. Note that their sample size in that study is relatively lower than the size that is generally required for artificial neural networks. Similarly, \cite{lingitz2018lead} applied some machine learning algorithms to get point estimates of manufacturing lead times for semiconductor manufacturing.

\cite{mori2015planning} conducted a comparison between Bayesian networks, ANN, and SVM to predict probability distributions of production loads and production
times for a plate mill process. They first classify production groups according to similar product features then computed the probability distribution for each production group. \cite{ozturk2006manufacturing} used a hypothetical manufacturing environment and synthetic data to test their decision tree approach augmented by an attribute selection scheme. \cite{pfeiffer2016manufacturing} used random forest for prediction and tested their approach on a synthetic data combined with a simulation based decision support system.

Our work has substantial differences with the prior work. Firstly, our uncertainly quantification is a novel approach. \cite{mori2015planning} used Bayesian networks before, however, they first grouped parts then fitted distributions to these groups. Whereas we output a distribution to each sample and employ  probabilistic models in more granular sense. Additionally, we propose and compare three state-of-the-art methods that have never been utilized for this problem and apply a post-hoc calibration method\citep{kuleshov2018accurate} to increase the accuracy of probabilistic outputs.

\section{Data}
\label{sec:data}

We exhibit our dataset in this section. We use a production dataset of a metal manufacturer that operates in Canada. The dataset consists material, thickness, cutting gas, part dimensions, sheet dimensions, cut dimensions, the number of holes, operation type, and rule-based lead time estimate of the manufacturer as input features. The rule-based lead time estimate is a natural baseline for regression metrics since the model we develop is going to replace that in production. In addition, we use it as an input since it can carry valuable information from the manufacturer's experience. By this way, we integrate domain knowledge into our models, which is known to improve models \citep{nabi2020bayesian}. For instance, according to the manufacturer bigger parts take more time or as the number of holes increase the compilation time takes longer. However, the exact relationship between the features and production time is unknown a priori and it presumably has non-linearity and feature interactions. Moreover, production times have idiosyncratic variation, as we explained in Section \ref{sec:intro}. To extract this relationship and quantify the uncertainty, we propose probabilistic models in this paper.\looseness=-1

\begin{figure}[]
\centering
\includegraphics[scale=0.4]{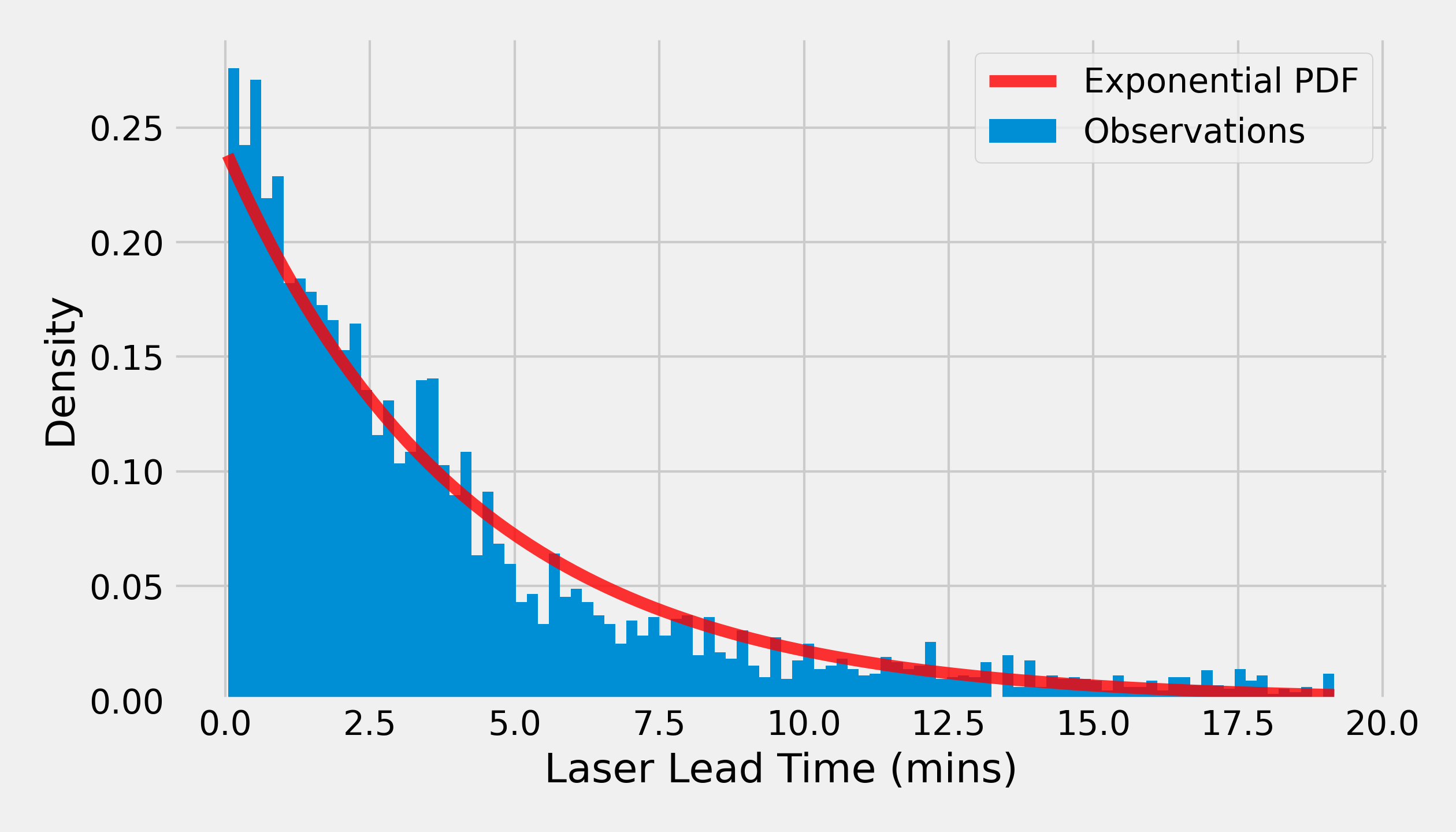}
\caption{The target variable can be estimated by exponential distribution with rate $\mu = 4.16$.} 
\label{fig:pdf}
\end{figure}

The output of the model is the realized production duration in the manufacturing facility. During our data validation we determined three problems in the dataset:
\begin{itemize}
    \item Zero values: Since a zero value is meaningless for production times, we asked for explanation of such values. The manufacturer confirms that these samples are erroneous. We disregard these samples.
    \item Extremely high values: 3\% of the data has greater than 25.35 minutes lead time (99\% quantile of $Exponential(4.16)$). We reached out the company for validation and reasons of such values. After an internal investigation, they determined that these values are caused by external(unrelated with process) factors. Despite we do not have explanatory ingredients for the group in the dataset, we decide to keep these values since they are not measurement errors.
    \item Small values: The 25\% quartile of the lead times is 1 minute. Without special treatment, this values are difficult to handle along with higher values. We explored $log$ transformation of the response variable. However, it degrades the overall model in this case. Thanks to the business objective, for the quick parts, we do not need the same level of accuracy that we need in the other part of the data. Hence, we did not apply a special treatment for small values.
\end{itemize}

After cleaning data errors and samples with missing values, the size of the dataset is 7841. 20\% of the dataset is used as the test set and 3-fold cross validation is used for the hyperparameter selection. Unless otherwise stated, all the results given in this paper are from the test set. It can be seen from Figure \ref{fig:pdf} that the laser lead times can be represented by a typical exponential distribution.

\section{Methods}

In this section, we present the algorithms that we used for lead time prediction. We develop four methods to have predictive intervals for the output instead of a point estimate. Three of our models are probabilistic models that are fitting a probability distribution as the output. The fourth model uses quantile regression to form predictive intervals.
\begin{enumerate}[label=\roman*.]
    \item \textbf{Probabilistic neural network (P-NN):} We build a fully connected neural network with a probabilistic output layer. The output of this model is Gaussian distribution and the training is done by maximizing the log-likelihood. The details about the model architecture can be found in the appendix.
    \item \textbf{Natural gradient boosting model (NGBoost):} NGBoost uses the natural gradient\citep{duan2020ngboost} to leverage a distribution as the output of the model in contrast with other tree-based boosting models. We optimized log-likelihood for exponential and Gaussian output distributions.
    \item \textbf{Gaussian process regression (GP) model:} Using a usual Gaussian process model is not feasible for large datasets since it has $O(N^3)$ computational complexity where $N$ is the size of the dataset. One way to deal with this complexity is variational inference \citep{hensman2013gaussian}. In this setting, the optimization is done by minimizing the evidence lower bound (ELBO).
    \item \textbf{Gradient boosted trees augmented by quantile regression (Q-GB):} Although this model does not output a distribution, it can be used to generate predictive intervals via quantile loss. We fit three gradient boosting models for upper, lower bounds of the predictive interval and mean/median predictions. According to our results, it works well in practice and the estimated prediction intervals are well calibrated. \looseness=-1
\end{enumerate}

\begin{table*}[]
\centering
\begin{tabular}{@{}lllll@{}}
\toprule
                                               & \multicolumn{2}{l}{Mean Predictions} & \multicolumn{2}{l}{Median Predictions} \\ \midrule
                                               & $R^2$            & MAPE              & $R^2$              & MAPE              \\ \midrule
Probabilistic Neural Network                   & 0.35             & 14.52             & \textbf{0.35}      & 14.52             \\
Natural Gradient Boosting (Exponential Output) & \textbf{0.40}    & \textbf{11.71}    & 0.32               & \textbf{8.07}     \\
Natural Gradient Boosting (Normal Output)      & 0.31             & 14.45             & 0.31               & 14.45             \\
Gaussian Process                               & 0.33             & 13.63             & 0.33               & 13.63             \\
Gradient Boosting with Quantile Regression     & 0.39             & 16.11             & 0.28               & 8.94              \\
Status Quo Predictions                         & 0.22             & 19.61             & N/A                & N/A     \\
Linear Regression                         & 0.35             & 13.72             & N/A                & N/A                 \\ \bottomrule
\end{tabular}
\caption{Model performance measures. Mean and median predictions differ for Natural Gradient Boosting and Gradient Boosting with Quantile Regression. Status quo predictions come from company estimates based on domain experience.}
\label{tab:results}
\end{table*}

The consistency between predicted and empirical probabilities is essential for probabilistic models \citep{simchi2020calibrating}. We report calibration performances of the models in Section \ref{sec:results}. We used isotonic regression technique proposed in \cite{kuleshov2018accurate} to carry out post-hoc calibrations for probabilistic outputs.

\section{Results}
\label{sec:results}

In this section, we compare the performances of the algorithms and benchmarks. The status quo benchmark is realized quotations of the company as explained in Section \ref{sec:data}. Additionally, we give linear regression results to show the improvement over a simple model. We use the coefficient of determination ($R^2$) and mean absolute percentage error (MAPE) as performance measures and calibration curves to evaluate probability calibrations. Using these metrics, we are able to compare all four algorithms, the status quo and linear regression benchmarks.

It can be seen from Table \ref{tab:results} that all of the models improve the company benchmark. NGBoost with exponential output is better than others in both mean and median predictions. Moreover, median predictions are better off in terms of MAPE and worse off in $R^2$. We note that the models whose output is normal distribution have the same mean and median. Since the target variable has an exponential distribution, the left skewness of exponential distribution makes an advantage for median predictions for MAPE. This has a reverse effect for $R^2$ since it normalizes the scale, therefore, predicting means is more meaningful for normalized values. 

The linear regression benchmark drastically improves the company status quo predictions. Among our candidate models, only NGBoost with Exponential output and Q-GB can improve the linear regression benchmark.

\begin{figure}[h!]
\centering
\includegraphics[scale=0.4]{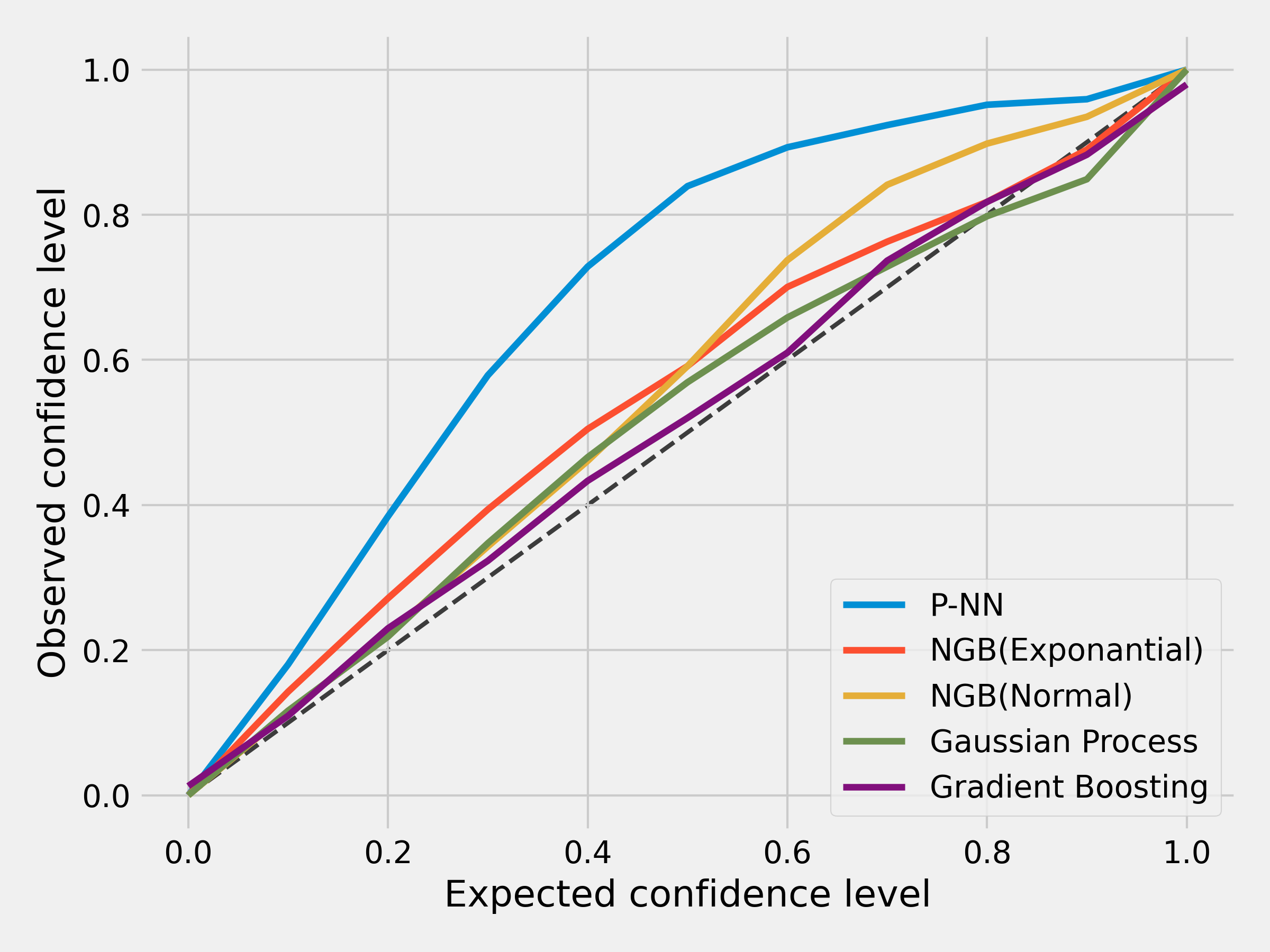}
\caption{Calibration curves of models. Perfect calibration is represented by the grey dashed line.} 
\label{fig:calib}
\end{figure}

Figure \ref{fig:calib} shows the alignment between predicted probabilities and empirical counterparts. The dashed line represents a perfect calibration. We give the results without a post-hoc calibration for brevity, post-hoc calibration results can be found in the Appendix.  All models have a tendency to have slight underconfidence. Although it is at negligible levels for the other models, one can observe that the P-NN is significantly underconfident. On the other hand, we can conclude that quantile regression augmented Gradient Boosting has the best calibration. 

The calibration success of quantile regression against other probabilistic models is remarkable. The learning process of Q-GB consists of separate point fits for the upperbound, mean, and the lowerbound. This makes it more accurate to determine the boundaries of the prediction interval. On the other hand, for a predetermined confidence, the Q-GB needs to be run three times to get the prediction interval and the mean. This extra computation is especially costly when the data size increases.

In our error analysis, we do not determine any insightful pattern of the error rates among different explanatory variable values and groups. We observe that among the three quartiles of the response variable, the first quartile which has the processes of less than 1 minute has worse metrics. In other words, the models fail to predict quick processes accurately. This is essentially expected for MAPE since it is sensitive to small values. We think that incorporating the information regarding the external factors mentioned in Section \ref{sec:data} can improve the performance for small values along with the overall performance. After our discussions with the manufacturer, we started collecting the data for any external interruptions during the manufacturing process. Using this information we plan to reduce the noise in the training dataset which will presumably increase the performance.

\section{Concluding remarks}

In this study, we applied probabilistic modelling techniques for the lead time prediction of metal manufacturing. Lead time prediction is essential for several business decisions and probabilistic modelling of this task is almost entirely ignored in practice. We fill this gap with four powerful methods and present our results in this paper. Our results show evident improvement over the company's current rule-based estimation along with well probability calibration. 

\nocite{malinin2020uncertainty}
\nocite{lim2019manufacturing}
\nocite{welsing2020combining}
\nocite{friedman2001elements}
\nocite{rahimi2020post}

\newpage
\bibliographystyle{plainnat}
\bibliography{references.bib}

\appendix

\section{Post-hoc Calibration Results}

\begin{figure}[h!]
\centering
\includegraphics[width=0.8\linewidth]{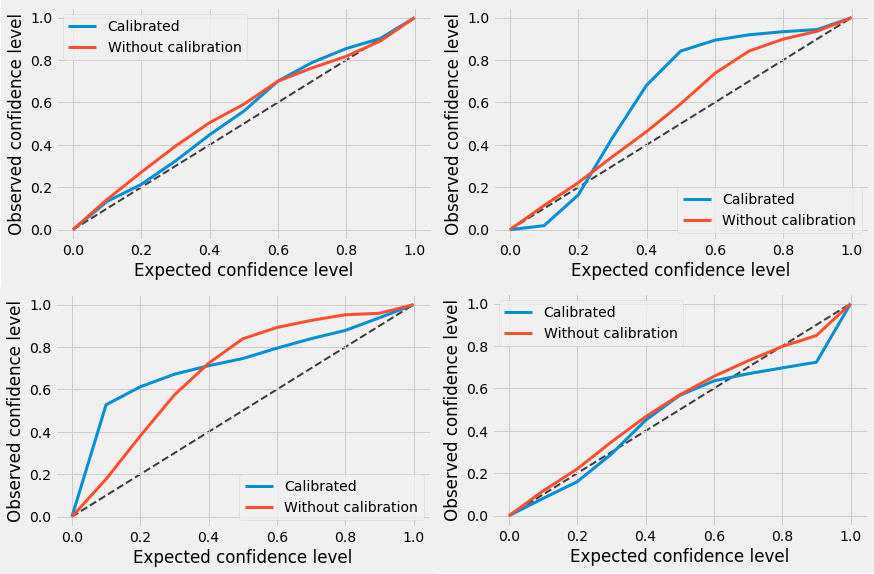}
\caption{Calibration curves of the raw outputs and calibrated outputs. Top left: NGBoost with Gaussian output. Top right: NGBoost with exponential output. Bottom left: Probabilistic neural network. Bottom right: Gaussian Processes.} 
\label{fig:calibs}
\end{figure}

While some of the models do not require an ad-hoc calibration, the method we used \citep{kuleshov2018accurate} failed to improve the models with bad calibration (e.g. NGBoost with Gaussian output and the probabilistic neural network).

\section{Model Parameters}

\textbf{Gaussian Processes.} We use the model proposed in \cite{hensman2013gaussian} and implemented using GPflow \citep{matthews2017gpflow}. We use ADAM\citep{kingma2014adam} to optimize the evidence lower bound with mini-batch size of 1000.

\textbf{Probabilistic Neural Network.} We use Gaussian distribution in the output layer and ADAM\citep{kingma2014adam} to optimize log-likelihood with learning rate $10^{-4}$. Our architecture consists of two hidden layers each with 256 nodes and L2 regularization factor of $0.01$. 

\textbf{NGBoost.} We use default hyperparameters for Gaussian and exponential output distribution.

\textbf{Quantile regression augmented Gradient Boosting.} We use Scikit-learn\citep{pedregosa2011scikit} implementation with the following hyperparameters:

\begin{table}[h!]
\centering
\begin{tabular}{@{}ll@{}}
\toprule
Hyperparameter      & Value \\ \midrule
learning\_rate      & 0.007 \\
max\_depth          & 13    \\
max\_features       & 3     \\
min\_samples\_leaf  & 5     \\
min\_samples\_split & 10    \\
n\_estimators       & 200   \\
subsample           & 0.65  \\ \bottomrule
\end{tabular}
\label{tab:hyperparams}
\end{table}

\end{document}